\documentclass[letterpaper, 10 pt, conference]{ieeeconf}  

\IEEEoverridecommandlockouts                              
                                                          
\usepackage{times} 
\usepackage{epsfig} 
\usepackage{graphics} 
\usepackage{amsmath} 
\usepackage{amssymb}  
\usepackage{mathptmx} 
\usepackage{xcolor}
\usepackage{arydshln}
\usepackage{algorithm,algorithmicx,algpseudocode}
\usepackage{gensymb}
\let\labelindent\relax
\usepackage{enumitem}
\usepackage{hyperref}
\usepackage{comment}

\title{\LARGE \bf
Deep Depth Estimation from Visual-Inertial SLAM}
\author{Kourosh Sartipi, Tien Do, Tong Ke, Khiem Vuong, and Stergios I. Roumeliotis$^{\dagger}$
\thanks{$^{\dagger}$Kourosh Sartipi, Tien Do, Tong Ke, Khiem Vuong, and Stergios I. Roumeliotis are with University of Minnesota, Minneapolis, MN 55455 {\tt\small sarti009@umn.edu, doxxx104@umn.edu, kexxx069@umn.edu, vuong067@umn.edu, stergios@umn.edu}.}
}

\begin{document}

\maketitle

\thispagestyle{plain}
\pagestyle{plain}

\begin{abstract}
This paper addresses the problem of learning to complete a scene's depth from sparse depth points and images of indoor scenes.
Specifically, we study the case in which the sparse depth is computed from a visual-inertial simultaneous localization and mapping (VI-SLAM) system.
The resulting point cloud has low density, it is noisy, and has non-uniform spatial distribution, as compared to the  input from active depth sensors, e.g., LiDAR or Kinect.
Since the VI-SLAM produces point clouds only over textured areas, we compensate for the missing depth of the low-texture surfaces by leveraging their planar structures and their surface normals which is an important intermediate representation.
The pre-trained surface normal network, however, suffers from large performance degradation when there is a significant difference in the viewing direction (especially the roll angle) of the test image as compared to the trained ones.
To address this limitation, we use the available gravity estimate from the VI-SLAM to warp the input image to the orientation prevailing in the training dataset.
This results in a significant performance gain for the surface normal estimate, and thus the dense depth estimates.
Finally, we show that our method outperforms other state-of-the-art approaches both on training (ScanNet~\cite{Dai2017scannet} and NYUv2~\cite{Silberman:ECCV12}) and testing (collected with Azure Kinect~\cite{KinectAzure}) datasets.
\end{abstract}

\section{Introduction}

Determining the dense depth of a scene has important applications in augmented reality, motion planning, and 3D mapping. 
This is often achieved by employing depth sensors such as Kinect and LiDAR. 
Besides the high-cost, size, and power requirements of depth sensors, their measurements are either sparse (LiDAR) or unreliable at glossy, reflective, and far-distance surfaces (Kinect).
Recent research has shown that some of the limitations of depth sensors can be overcome by employing RGB images along with the strong contextual priors learned from large-scale datasets ($\simeq$ 200K images), using a deep convolutional neural network (CNN).
Specifically, to produce dense depth estimates, recent approaches have employed CNNs with three types of inputs: (a) a single RGB image (e.g.,~\cite{eigen2014depth, fu2018dorn, laina2016DeeperDepthPrediction, liu2015learning, li2018megadepth});
%
(b) poses (position and orientation) and optical flow from multiple images (e.g.,~\cite{liu2019neuralrgbd, ummenhofer2017demon, zhou2018deeptam});
(c) a single RGB image and sparse depth~\cite{tang2019sparse2dense, liao2017parse, cheng2018cspn, Mal2018SparseToDense, qiu2019deeplidar, yang2019DDP, teixeira2020aerial_dc, wong2020dc_vio, li2019ManneqinChallenge, zhang2018depthcompletion}.

In our work, which falls under (c) and is inspired by~\cite{qiu2019deeplidar}, we seek to estimate dense depth by fusing RGB images, learned surface normals, and sparse depth information. 
In contrast to~\cite{qiu2019deeplidar} that uses LiDAR, we obtain the sparse points from a real-time visual-inertial SLAM (VI-SLAM) system~\cite{KejianWuInverseFilter}. 
There are thee key differences between the point clouds directly measured by a Kinect or a LiDAR and those estimated by VI-SLAM: Density, accuracy, and spatial distribution.
In particular, the VI-SLAM point cloud comprises sparse points ($\simeq$ 0.5\% of an image’s pixels) that are extracted and tracked across images and triangulated using the camera's estimated poses. 
For this reason, the accuracy of these points varies widely, depending on the local geometry and the camera's motion. 
Moreover, the VI-SLAM points are not uniformly distributed across an image.
They are usually found on high-texture surfaces, while textureless areas such as walls, floors, and ceilings that are ubiquitous in man-made environments often contain significantly fewer points. 
Due to this large domain gap, networks trained using Kinect or LiDAR data suffer from a significant performance degradation when
given 3D points triangulated by VI-SLAM as sparse depth information~\cite{teixeira2020aerial_dc, wong2020dc_vio}.
Furthermore, in large-scale indoor datasets (ScanNet~\cite{Dai2017scannet}, NYUv2~\cite{Silberman:ECCV12}, Matterport3D~\cite{Matterport3D}), the images are usually aligned with gravity~\cite{saito2020roll_depth}. 
During inference time, this bias results in further performance degradation of the depth completion networks~\cite{cheng2018cspn}, as well as optical-flow to depth networks trained on these data~\cite{liu2019neuralrgbd}. 
%

A straightforward approach to address the domain gap and the lack of images from various vantage points is to collect and process more data and re-train the network. 
This, however, is both labor and time intensive. 
Alternatively, 3D mesh reconstructions from RGB-D sequences (e.g., BundleFusion~\cite{dai2017bundlefusion}, BAD-SLAM~\cite{schops2019badslam}, etc.) can be employed to synthesize novel views of RGB images. 
The accuracy of the 3D mesh, however, is usually not sufficient and the quality of the resulting data depends on many factors, e.g., the overlap between frames, the sparse points tracking error, etc. 
For this reason, in our work, we employ the VI-SLAM point cloud when training the depth estimation network while incorporating the VI-SLAM estimate of gravity direction to reduce the effect of ``unseen orientation."
Specifically, to address the domain gap issue in the point cloud, we first train a network using sparse depth input from the point cloud generated by VI-SLAM, instead of randomly sampling from the ground-truth depth (e.g., \cite{cheng2018cspn}). 
Furthermore, to increase the density of the point cloud, we leverage the planar structures commonly found indoors. 
Secondly, to solve the domain gap in viewing directions, we align each input image taken with ``unseen orientation" to a rectified orientation that the pre-trained network is more familiar with, using the gravity direction estimated from VI-SLAM.
We show that this approach not only improves the performance on images taken with familiar orientations, but also achieves satisfactory generalization on the unfamiliar ones.
In summary, our main contributions are:
\begin{itemize}
    \item We introduce an efficient approach to improve the generalization of the VI-SLAM depth completion that leverages (i) the planar geometry of the scene and (ii) the camera's orientation with respect to gravity.
    \item We implement a full pipeline from VI-SLAM to dense depth estimation for evaluation on Azure Kinect~\cite{KinectAzure} and perform extensive experiments that demonstrate the advantages of our method over state-of-the-art approaches on dense-depth estimation~\cite{liu2019neuralrgbd, cheng2018cspn}.
\end{itemize}

\section{Related Work}
\textbf{Single-view depth estimation}. Depth estimation from a single image has been studied by early works such as~\cite{saxena2006learning} which is based on handcrafted features.
More recently, numerous deep learning-based approaches have appeared (e.g., \cite{fu2018dorn,eigen2014depth,laina2016DeeperDepthPrediction, liu2015learning}) for estimating dense depth.
Note, however, that given a single image, the pixels' depths cannot be determined only from local features; hence data-driven approaches have to rely on the global context of the image, which is learned from the training data.
Therefore, despite the surprisingly good performance when trained and tested with images from the same dataset, they exhibit poor generalization on cross-dataset experiments~\cite{li2018megadepth}.

\textbf{Multi-view depth estimation}. One way to overcome the single-view depth estimation challenges is to consider multiple camera poses and optical flow.
Specifically, the scene depth can be recovered up to scale, given information from multiple views, or with metric scale if the poses are estimated with the aid of other sensors (e.g., IMU).
In particular, \cite{ummenhofer2017demon} takes two images as input, estimates the optical flow as an intermediate result, and refines the depths as well as the poses iteratively. 
On the other hand, \cite{zhou2018deeptam} computes the photometric errors by warping adjacent images to the current one, and inputs them along with the current image to neural network.
Lastly, \cite{liu2019neuralrgbd} estimates a probability distribution for the depth, instead of a single depth, and refines the initial depth estimates from a neural network by warping the depth distribution of adjacent images and fusing them in a Bayesian fashion.
These methods implicitly estimate depth from poses and optical flow using exclusively neural networks, instead of computing the depths of at least some points directly from geometry. As shown in~\cite{li2019ManneqinChallenge}, employing these sparse or semi-dense depths as the input of neural network results in higher performance as compared to relying on the optical flow.

\textbf{Depth completion from RGB and sparse depth}.
A key difference between the methods described hereafter and the previous two families of approaches is that the domains of their inputs are significantly different. 
Specifically, the RGB image has a well-defined range of values for all pixels, while the sparse depth image has only few valid values where the majority of the pixels' depths are unknown. 
To address this domain difference, many approaches such as \cite{eldesokey2019confidence, huang2019hms} propose normalized convolution and upsampling layers so that the missing pixels will not be processed in the convolution kernels.
%
%
On the other hand, methods such as \cite{jaritz2018sparse, van2019sparse,Mal2018SparseToDense, ma2019selfsupervised, lee2019depth} do not treat the sparse normal input differently, which indicates with proper training applying standard convolution can achieve comparable results.
%

A different approach proposed by \cite{qiu2019deeplidar} showed that employing a network to compute the surface normals based on the RGB image and using the normals as an additional input to the later layers can improve the network's performance.
Moreover, it was proposed to concatenate color/normal channels and sum the depth channels for the skip-connections between the encoder and decoder.
As we show in our ablation study, while this design choice results in improvement on the same scenes, it reduces the generalizability to, e.g., data collected from different devices.

In~\cite{cheng2018cspn}, Cheng et al. introduced the convolutional spatial propagation network (CSPN), where the initial depth image and an affinity matrix are first produced by a CNN and then the depth is iteratively refined through a diffusion process involving the affinity matrix and the current depth estimate.
Furthermore, for the case of depth completion, CSPN employs validity masks to retain the depth of the sparse input, and hence does not allow the network to correct potentially noisy measurements.
As we will show in our experimental results, this policy will lead to loss in accuracy when the input depth is noisy.

Closely related to our work are those of \cite{teixeira2020aerial_dc} and \cite{wong2019voiced}.
In \cite{teixeira2020aerial_dc}, a system to compute dense depth from either LiDAR or SLAM point-cloud on-board MAVs is described using the confidence propagation network similar to \cite{eldesokey2019confidence}.
Since, however, their focus is on real-time performance on devices with limited resources, their accuracy is lower than the state of the art.
As in our work, \cite{wong2019voiced} also employs a VI-SLAM system to generate the sparse input. Moreover, it uses a two-step process to first extend the depth data from the sparse 3D point cloud and then combine them with the RGB image as the input to a network that produces the final dense-depth image.
In particular, \cite{wong2019voiced} first employs Delaunay triangulation~\cite{barber1996quickhull} to fit a triangular mesh on the sparse points of the image and then computes the depth of all the pixels falling within each triangle using its plane equation. 
Note that the accuracy of this approach depends heavily on the assumption that the triangles match with the planar surfaces of the image's scene, which, in general, is not true.\footnote{Another contribution of \cite{wong2019voiced} is creating visual-inertial datasets for depth estimation. However, it does not have ground-truth surface normal, which our networks relies on. Therefore, we instead employ Scannet~\cite{Dai2017scannet}, with surface normal readily available from 3D mesh, to train our network. As for evaluating generalization capability, we collected our own dataset with sufficient roll and pitch variation (which is not the case for the dataset of~\cite{wong2019voiced}) to highlight the effect of incorporating gravity.}

\begin{figure*}[ht]
\includegraphics[width=\textwidth]{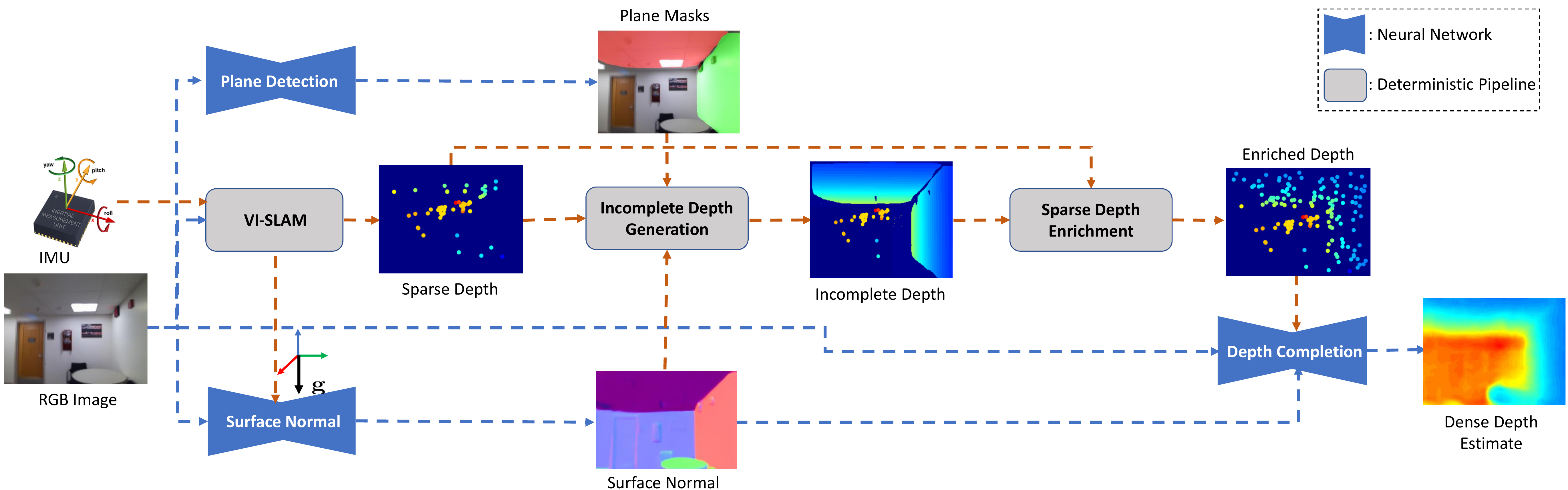} 
\centering
\caption{Overview of the system. At each keyframe, the VI-SLAM (Sec.~\ref{ssec:slam}) processes images and IMU measurements to compute (i) the gravity direction which is provided to the surface-normal network (Sec.~\ref{ssec:normal}), and (ii) the sparse depth image. The sparse depth enrichment module (Sec.~\ref{ssec:partial_depth}) increases the density of the sparse-depth image using information about planes in the image. This is provided by the plane detection network (Sec.~\ref{ssec:mask}) that classifies image pixels as belonging to a particular plane and the surface normal network that estimates the normal of each pixel. The depth completion network (DCN) (Sec.~\ref{ssec:infer_depth}) employs the RGB image, the pixel normals, and the enriched sparse-depth image to produce a dense depth image.}
\label{fig:overview}
\end{figure*}

\section{Technical Approach}\label{sec:method}
In this paper, we propose a method to accurately predict dense depths using only sensors available on most mobile devices, i.e., a camera and an IMU.
Fig.~\ref{fig:overview} depicts an overview of our system, where the depth completion network (DCN) estimates the dense depth from the following inputs:
(i) the RGB image, (ii) a sparse depth image based on the 3D points triangulated by the VI-SLAM, and (iii) the
surface normal map predicted by another CNN. As shown experimentally by~\cite{qiu2019deeplidar} and evident in our ablation studies, (iii) improves the accuracy of depth estimation. 

Specifically, we employ VI-SLAM~\cite{KejianWuInverseFilter} (see Sec.~\ref{ssec:slam}), which takes as input images and IMU measurements, to compute the \emph{sparse 3D point cloud} observed by the camera.
Then, the \emph{sparse depth image} is obtained by projecting the point cloud to the camera frame.
Additionally, a CNN predicts the \emph{surface normal} of every pixel in the RGB image (see Sec.~\ref{ssec:normal}) while leveraging the gravity direction estimated by the VI-SLAM to improve its accuracy.
Note that although the sparse depth image from the VI-SLAM can be used directly as input to the DCN, we seek to first increase its density by performing a sparse-depth enrichment step.
To do so, we extract the \emph{planar patches} of the image using a CNN (see Sec.~\ref{ssec:mask}), and use the estimated normals along with any 3D points that fall inside a plane, to compute a denser depth representation of the scene (see Sec.~\ref{ssec:partial_depth}).
Finally, the DCN (see Sec.~\ref{ssec:infer_depth}) computes the dense depth estimate based on the RGB image, the enriched sparse depth image, and the surface normals.

\subsection{VI-SLAM: Sparse Depth Image Generation}\label{ssec:slam}
In this work, we employ the inverse square-root sliding window filter (SR-ISWF)~\cite{KejianWuInverseFilter} to estimate in real-time camera poses and 3D feature positions.
Specifically, at each time step the SR-ISWF extracts FAST~\cite{FAST} corners in the current image, and tracks them by matching their corresponding ORB descriptors~\cite{ORB} to the previous images.
The SR-ISWF then fuses the IMU measurements and the 2D-to-2D feature tracks across the sliding window to estimate the camera's motion along with the features' positions.

Every time the SR-ISWF processes an image, we project all visible 3D features on the image and use their depth (i.e., their Z component) to create the sparse depth image. The SR-ISWF also computes the gravity direction which is passed to the surface normal network.
%
%

\subsection{Surface Normal Network}\label{ssec:normal}

%

\begin{figure}[ht]
\includegraphics[width=0.5\textwidth]{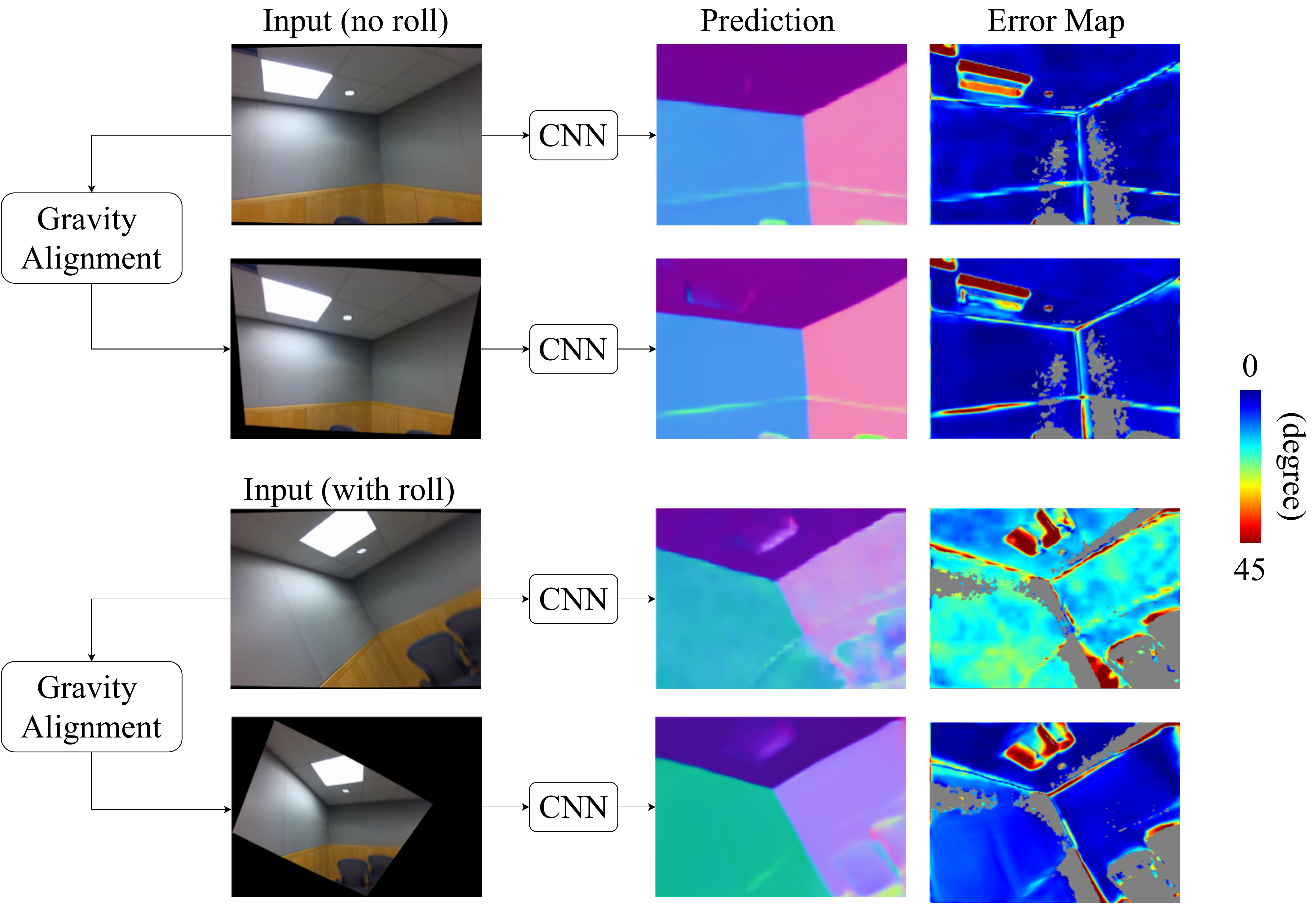} 
\centering
\vspace{-5mm}
\caption{Top row: Input image from Azure Kinect along with the gravity aligned image; the performance of FrameNet is satisfactory. Bottom row: Input image with significant roll component, resulting in poor performance. Warping the input based on the gravity direction before passing it through the CNN improves the performance in both cases.}
\label{fig:surface_normal_error_with_roll}
\end{figure}

\begin{figure*}[ht]
\includegraphics[width=\textwidth]{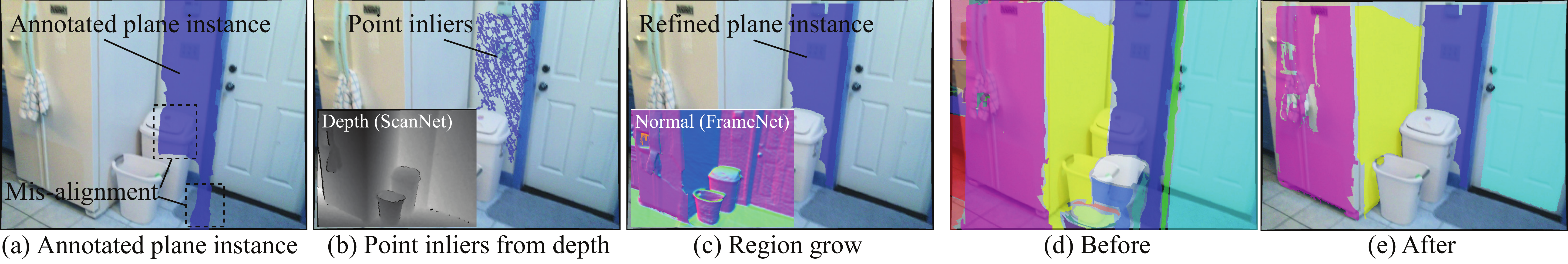} 
\centering
\vspace{-7mm}
\caption{We refine the imprecise plane annotation: (a) Given an imprecise plane instance, (b) we fit a plane using the depth from ScanNet with a strict threshold, which produces a set of sparse, yet accurate, points. (c) Using the point set, we incrementally grow the planar region based on the surface normal from FrameNet. (d-e) This refinement allows us to precisely label the plane instances.}
\label{fig:annotation_cleanup}
\end{figure*}

As it will become evident from our experimental validation, the performance of the DCN depends on the accuracy of the surface normals prediction. In particular, training state-of-the-art surface-normal-estimation networks such as the FrameNet~\cite{Huang19FrameNet} on large-scale indoor datasets (e.g., ScanNet~\cite{Dai2017scannet}, NYUv2~\cite{Silberman:ECCV12}, Matterport3D~\cite{Matterport3D}) does not yield satisfactory results. 
This is due to the fact that the FrameNet's surface-normal estimator's performance degrades significantly when tested on images whose roll angles deviate substantially from the vertically-aligned images used during training (see Fig.~\ref{fig:surface_normal_error_with_roll}).
%

To address this issue, we follow the approach of~\cite{Do2020SurfaceNormal} to warp the input image so that the gravity, which is estimated from IMU, is aligned with the image's vertical axis.
Note that, in this paper, we are interested in how the accuracy of surface normal prediction affects the depth completion performance, rather than solely focusing on the surface normal performance as in~\cite{Do2020SurfaceNormal}.
%
%

The idea of using gravity from VI-SLAM has been proposed in~\cite{fei2019geosupervised} as a regularization for a depth-prediction network during \textit{training} time.
In contrast, in our work we employ the online estimated gravity to improve the generalization during \textit{inference} time.
Recently, \cite{saito2020roll_depth} proposed to remove the roll angle from an image by using the orientation estimated from monocular SLAM~\cite{ORBSLAM}. This method, though, lacks global information hence it relies on the assumption that the first camera's view is aligned with gravity.
In contrast, and due to the observability of gravity in VI-SLAM system~\cite{Hesch_TRO_14}, our method makes no assumptions about the camera's motion.

\subsection{Plane Detection}\label{ssec:mask}
A key idea behind our approach is to take advantage of planar surfaces present in the scene to enrich the sparse depth image (see Sec.~\ref{ssec:partial_depth}).
To do so, we predict plane masks on the image using a CNN. 
Specifically, we leverage the ground-truth plane masks from Plane-RCNN~\cite{Liu2019planercnn} which employs the 3D mesh reconstruction created from ScanNet~\cite{Dai2017scannet} for the multi-plane instance proposals.
Unfortunately, the resulting plane annotations are misaligned as shown in Fig.~\ref{fig:annotation_cleanup}. 
Hence, to obtain reliable training data, Plane-RCNN proposed a heuristic method to detect this misalignment based on the discrepancies between the projected 3D mesh reconstruction on each image and their corresponding depth, and then drop any frames with large discrepancies during training.
While effective, this method discards a large set of planes annotations during training, which can potentially degrade the plane detector's performance.

To address this problem, we employ RANSAC-based~\cite{Fischler1981ransac} plane fitting and region growing to refine the plane annotations. 
Specifically, for each annotated plane from Plane-RCNN, we employ 3pt RANSAC, using the corresponding depth  from ScanNet and a strict inlier threshold (2 cm) to fit a plane.
This yields only a small set of inliers due to the imprecise depth measurements. 
Given the initial inliers, we grow the coplanar region through neighboring pixels based on two criteria: (i) the distance of each 3D point to the plane is less than 20 cm and (ii) the corresponding surface normal from FrameNet~\cite{Huang19FrameNet} is close (less than 30$^{\circ}$) to the plane's normal. 
If the plane computed through this process is substantially smaller than the annotation, we discard the annotation. 
By applying this method to all the annotated planes we are able to accurately compute the plane instances (see Fig.~\ref{fig:annotation_cleanup}) and ensure that the planes are well-aligned with the RGB images.
Lastly, we employ the improved annotations as ground-truth to train a Mask~R-CNN~\cite{he2017maskrcnn} network for plane detection.


\subsection{Sparse Depth Enrichment}\label{ssec:partial_depth}

In this module, we enrich the sparse depth image by increasing the number of points it contains.
We focus on the parts of the scene where 3D point features project on a detected plane (see Sect.~\ref{ssec:mask}).
This process comprises the following two steps for estimating each plane's parameters:

\begin{enumerate}
    \item Plane Normal estimation: We randomly select a few pixels as plane-normal hypotheses and compare their directions to the rest of the plane's pixels to find the largest set of normal directions aligned within 10 degrees. We then assign as the plane's normal the average of the normal vectors of the largest set.
    \item Plane Distance estimation: Given the plane's normal $\mathbf{n}$, each 3D point $\mathbf{p}_i$ expressed in the camera's coordinate frame is a candidate for computing the plane's distance hypothesis $d_i = -\mathbf{n}^T\mathbf{p}_i$. As before, the one with the largest set of inliers is accepted, and the plane's distance $d$ is set to be the average distance of inliers.
\end{enumerate}
Next, we employ $\mathbf{n}$ and $d$ to compute the depth $z_i = -\frac{d}{\mathbf{n}^T \mathbf{b}_i}$ of each pixel, where $\mathbf{b}_i$ is its normalized homogeneous coordinate.
%
%
%
%
%
Hereafter, we refer to the depth image resulting from this process as the \textit{``incomplete depth image''} (see Fig.~\ref{fig:overview}).

Once the incomplete depth image is generated, we select only a subset of the points from it to form the enriched depth image provided to the DCN (see Fig.~\ref{fig:overview}).
The reason behind this choice is that often the 3D points triangulated by VI-SLAM may contain errors.
In this case, using the entire incomplete depth image as the input to the DCN will bias its output by implicitly forcing it to trust the depth values of many densely distributed pixels with erroneous estimates.
%
On the other hand, our experiments (see Sect.~\ref{ssec:nyu}) and those of~\cite{teixeira2020aerial_dc} show that the DCN performs significantly better when the sparse depth image is uniformly distributed over the RGB image.
This requirement, however, is not typically satisfied by indoor areas containing large textureless surfaces.
Instead, 3D points often cluster in few parts of an image where high texture is observed.
%
%
By employing the proposed enrichment procedure, however, we are able to reduce the gap between the initial distribution of sparse depth and a uniform distribution.
As shown in Sect.~\ref{sec:exp_results}, sampling between 100-200 points from the incomplete depth image produces the best results.

\subsection{Depth Completion Network}\label{ssec:infer_depth}
\begin{figure}[t]
\includegraphics[width=0.5\textwidth]{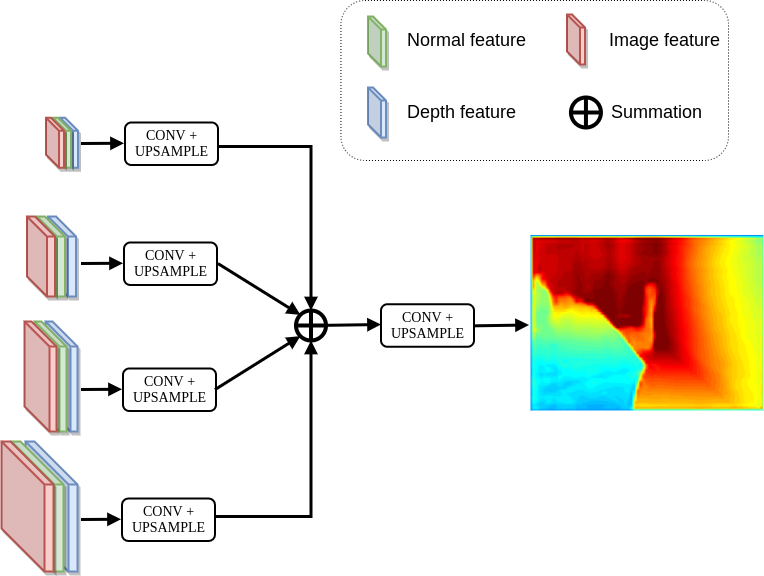} 
\centering
\caption{Decoder structure. Features at different scales of the RGB, normal, and depth images are concatenated together, passed through convolution and upsampling, and then summed. A final convolution and upsampling is applied to compute the dense depth image.}
\label{fig:deep_comp}
\end{figure}
Our DCN is inspired by the Panoptic FPN~\cite{kirillov2019panoptic}.
Specifically, each of the RGB, normal, and the sparse depth images are processed through separate encoders, which compute four feature tensors per input corresponding to different layers of the ResNet.
%
Before being processed by the decoder, we concatenate the features of the RGB, normal, and sparse depth images at each scale and apply convolution and upsampling, resulting in four feature tensors of size $128\times 60 \times 80$.
These are then summed together and passed through a final convolution and upsampling layer resulting in the dense-depth image (see Fig.~\ref{fig:deep_comp}).
%


\section{Experimental Results}\label{sec:exp_results}
In what follows, we experimentally verify the performance of our method, and compare it to state-of-the-art approaches.
Although our target application is for VI-SLAM systems, the lack of visual-inertial training data has led us to employ RGB-D datasets (i.e., ScanNet~\cite{Dai2017scannet} and NYUv2~\cite{Silberman:ECCV12}) to train our networks.
To assess the performance on visual-inertial data and verify the generalization capability of our approach, we have also collected data using Azure Kinect~\cite{KinectAzure}.
In terms of depth error metrics, we report root mean square error (RMSE) in meters, and $E(\hat{D}, \delta)$, which specifies the percentage of the estimated depths $\hat{D}$ for which $\max(\frac{\hat{D}}{D},\frac{D}{\hat{D}}) < \delta$, where $D$ is the ground-truth depth.
For surface normal error metrics, we report mean absolute of the error (MAD), median of absolute error (Median), and the percentage of pixels with angular error below a threshold $\xi$ with $\xi = $ 11.25\degree, 22.5\degree, 30.0\degree.



\textbf{Experiment Setup}: All the networks in this paper have been implemented in PyTorch~\cite{pytorch}, and the original authors' code and provided network weights have been used for comparisons against other methods.
To train the DCN, we employ $L_1$ loss and the Adam optimizer~\cite{kingma2015adam} with a learning rate of $10^{-4}$.
We train the model for 20 epochs and report the best epoch on the corresponding dataset's validation set.
The training was done on an NVIDIA Tesla V100 GPU with 32GB of memory with a batch size of 16.
%
%
%
Since the aspect ratio of the Azure Kinect dataset images is significantly different than those of ScanNet and NYUv2, we first crop them and then resize to $320\times 240$.
For the ScanNet and NYUv2 we only apply resizing.
Our neural network code is available at \url{https://github.com/MARSLab-UMN/vi_depth_completion} along with our datasets.

\subsection{Comparison on ScanNet Datasets}\label{ssec:exp_scannet}
We train and evaluate different configurations of our approach on the ScanNet indoor datasets and compare our performance with NeuralRGBD~\cite{liu2019neuralrgbd}.
Although ScanNet does not contain inertial data, we leverage other available annotations to estimate (i) the gravity direction, and (ii) a 3D point cloud, required for our training and testing.
Specifically, we obtain the gravity direction from the normal direction of pixels labeled as the ground (by FrameNet~\cite{Huang19FrameNet}).
To compute the 3D point cloud, we first extract FAST corners~\cite{FAST}, track them via KLT~\cite{Kanade1981}, and remove outliers by employing the 5pt-RANSAC~\cite{Nister04}.
The inlier tracks are then triangulated using the provided camera poses to generate the sparse 3D point cloud (on average, for each image we compute 58 sparse depth values from the point cloud which corresponds to $0.07\%$ of pixels).
Lastly, we generate plane masks for the entire dataset and compute the incomplete depth images through the process described in Sect.~\ref{ssec:partial_depth}.
%
%
%
%

%
%
%
%
%
%
%
%
In Table~\ref{tab:scannet}, we evaluate the accuracy of the reconstructed sparse and incomplete depth inputs as well as the final depth output using the following configurations: 
%
\begin{itemize}
    \item \emph{Triangulation}: The sparse depth resulting from projecting the 3D point cloud on the image.
    \item \emph{Incomplete Depth}: The incomplete depth images generated through the process described in Sect.~\ref{ssec:partial_depth}.
    \item \emph{NeuralRGBD}~\cite{liu2019neuralrgbd}: State-of-the-art in depth estimation from a sequence of images.
    \item \emph{Ours-SD}: DCN results trained and tested using the sparse depth as input.
    \item \emph{Ours-ID}: DCN results trained and tested using the incomplete depth as input.
    \item \emph{Ours-Enriched 100, 200}: Using \emph{Ours-SD} trained network with selecting 100, or 200 additional points from the incomplete depth to generate the enriched depth (see Sect.~\ref{ssec:partial_depth} and Fig.~\ref{fig:overview}).
\end{itemize}
%

As evident from Table~\ref{tab:scannet}, the error $E(\hat{D}, \delta)$ of the triangulated points and incomplete depths show that the inputs are relatively accurate.
%
Furthermore, using the enriched depth with 100 points as input has slightly higher accuracy in the stricter metrics ($\delta=1.05,1.10$). 
On the other hand, a denser enriched depth image comprising 200 points decreases the accuracy for the reasons explained in Sect.~\ref{ssec:partial_depth}. 
Although the network trained and tested using incomplete depth performs slightly better than both sparse and enriched depths for the ScanNet dataset, its performance degrades drastically on cross-dataset (see Sect.~\ref{ssec:exp_azure}).
%
%
%
%
Finally, these results show that our approach outperforms NeuralRGBD on all metrics, especially the stricter ones. 
%

\begin{table}[t]
\caption{Performance of depth completion on ScanNet test set}
\vspace{-5mm}
\label{tab:scannet}
\begin{center}
\resizebox{\columnwidth}{!}{
\begin{tabular}{|c|c|ccccc|}
\hline
 &                 &                   &                   &   $E(\hat{D}, \delta)$       &                     &  \\ \cline{3-7}
 &	RMSE $\downarrow$ & 1.05 $\uparrow$ &	1.10$\uparrow$  &	1.25$\uparrow$ &	$1.25^2 \uparrow$ &	$1.25^3 \uparrow$ \\
\hline
\hline
Triangulation            & 0.153 & 75.89 & 92.19 & 98.51 & 99.64 & 99.89 \\
Incomplete Depth               & 0.201 & 66.39 & 85.30 & 96.64 & 99.05 & 99.60 \\
\hline
NeuralRGBD~\cite{liu2019neuralrgbd}          &	0.294         &	46.35          &	72.99          &	93.00         &	98.30 &	99.43 \\

Ours-SD &	0.266 &	54.09 &	78.41 &	94.68 &	\textbf{98.91} &	\textbf{99.71} \\
Ours-ID & \textbf{0.265} &	54.65       &	   \textbf{78.71}       & \textbf{94.72}	 & 98.85	 &	99.67 \\
Ours-Enriched 100 &  0.271 & \textbf{54.85} & 78.54 & 94.55 & 98.76 & 99.66  \\
Ours-Enriched 200  & 0.271 &	54.37 &	78.11 &	94.46 &	98.73 &	99.63 \\
\hline
\end{tabular}
}
\end{center}
\vspace{-5mm}
\end{table}
\subsection{Comparison on NYUv2 Depth Dataset}\label{ssec:nyu}
To compare against the CSPN~\cite{cheng2018cspn}, we also test our approach on the NYUv2 datasets, comprised of color and depth images from 464 different indoor scenes acquired by a Microsoft Kinect sensor.
In our evaluation, we used the official 249 training scenes and sub-sampled $25,000$ color-depth image pairs for the network to train, while tested on the standard 654-image test set.
%
%
%
%
Since NYUv2 does not provide camera poses, to generate the sparse depth, we first extract 2D-to-2D correspondences (see Sect.~\ref{ssec:exp_scannet}), and then sample from the ground-truth depth instead of projecting a 3D point-cloud as input to the DCN.
%
%
%
%

Table~\ref{tab:nyu} shows the performance of the proposed approach compared to CSPN using their pre-trained model.
Note that the CSPN model has been trained on sparse depth generated by randomly sampling from the ground-truth depth.
In our tests, we compare the computed depth when the sparse depth is from 2D-to-2D tracks (denoted as ``Feature'' in Table~\ref{tab:nyu}), and by randomly sampling 200 points from the ground-truth (``Uniform'').
%
%
These results show that our approach outperforms CSPN in all metrics.
Additionally, the gap in performance when employing features is much larger than using random samples.
This is due to the distribution of FAST corners on the images, where more points are located on the textured areas as compared to the textureless ones.
%

%
%
\begin{table}[t]
\caption{Performance of depth completion on NYUv2 test set}
\vspace{-5mm}
\label{tab:nyu}
\begin{center}
\resizebox{\columnwidth}{!}{
\begin{tabular}{|c|c|ccccc|}
\hline
 &                 &                   &                   &   $E(\hat{D}, \delta)$       &                     &  \\ \cline{3-7}
 &	RMSE $\downarrow$ &	1.05 $\uparrow$ &	1.10$\uparrow$  &	1.25$\uparrow$ &	$1.25^2 \uparrow$ &	$1.25^3 \uparrow$ \\
\hline
\hline
CSPN\cite{cheng2018cspn} (Feature)   & 0.46  & 69.31 & 78.39 & 86.47 & 91.65 & 94.57 \\
Ours-SD (Feature) & \textbf{0.20}  & \textbf{80.38} & \textbf{91.27} & \textbf{97.53} & \textbf{99.50} & \textbf{99.89} \\
\hline
CSPN\cite{cheng2018cspn} (Uniform)   & \textbf{0.18} & 87.56 & 94.17 & 98.24 & 99.59 & 99.88 \\
Ours-SD (Uniform) & \textbf{0.18} & \textbf{88.86} & \textbf{94.82} & \textbf{98.46} & \textbf{99.66} & \textbf{99.91}  \\
\hline
\end{tabular}
}
\end{center}
\vspace{-5mm}
\end{table}
\subsection{Generalization Capability on Azure Kinect Datasets}\label{ssec:exp_azure}
In order to evaluate our DCN on datasets with visual-inertial data, and to better compare the generalization capability of our approach against the alternative ones, we collected 24 datasets in indoor areas using the Azure Kinect~\cite{KinectAzure}. 
Each dataset comprises color and depth images at 30~Hz, as well as IMU measurements at 1.6~KHz. 
The depth images are employed as the ground-truth, while the color images and the IMU measurements are processed by the VI-SLAM to compute the camera's poses and the triangulated feature positions.
These datasets produce a total of $\sim$8K keyframes which we use for evaluation.
Unless otherwise specified, all networks considered in this section are trained only on ScanNet dataset.

First, we present the performance and the effectiveness of our gravity alignment for the surface normal estimation (Sec.~\ref{ssec:normal}) on Azure Kinect datasets. 
%
%
Table~\ref{tab:surface_normal_azure} compares the proposed gravity alignment (see Sect.~\ref{ssec:normal}) against the following alternatives: (i) Vanilla - training with standard ground-truth surface normal and (ii) Warping augmentation - improving generalization by warping the input image with random rotation during training; on two categories of images (1) gravity aligned and (2) gravity non-aligned.
The network used in all cases is DORN~\cite{fu2018dorn}, and is trained with the truncated angular loss~\cite{Do2020SurfaceNormal}.
In these tests, we achieve satisfactory performance when evaluating DORN on the scenes with gravity aligned images (see Table~\ref{tab:surface_normal_azure}, Gravity-aligned frames).
Its performance, however, reduces drastically when the images have large pitch/roll rotations (see Table~\ref{tab:surface_normal_azure}, Gravity-nonaligned frames).
We attribute this degradation to the lack of training images with large pitch/roll angles. 
To assess the accuracy of our approach, we also compare against a naive data augmentation scheme that randomly warps each input image to provide the network with more diverse viewing-directions.
%
%
\begin{figure*}[h]
\includegraphics[width=\textwidth]{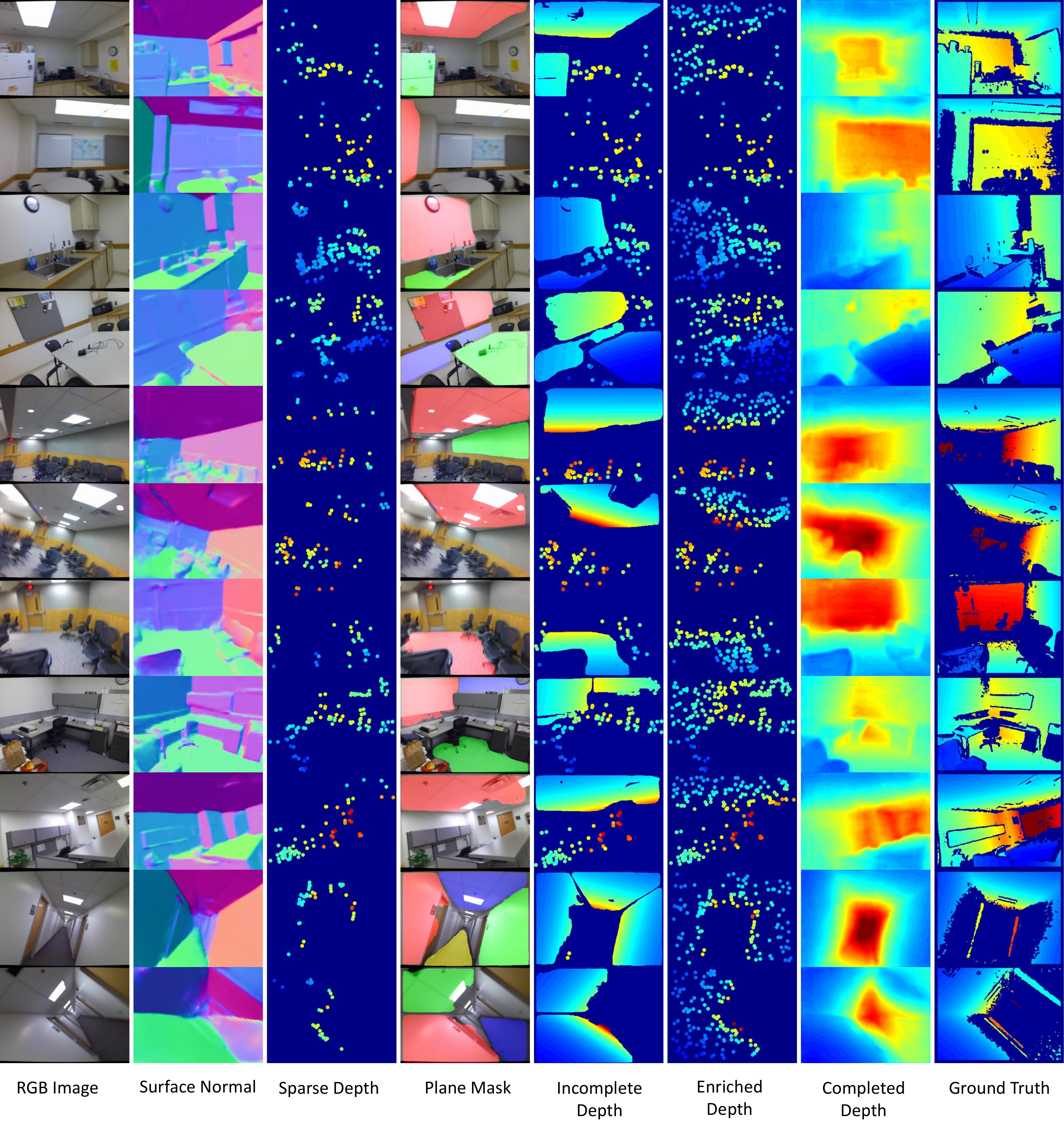} 
\centering
\vspace{-10mm}
\caption{Qualitative results on some Azure Kinect scenes.}
\label{fig:qualitative}
\end{figure*}
Finally, Table~\ref{tab:surface_normal_azure} shows that by warping in a direction aligned with the gravity obtained from VI-SLAM, our method outperforms both the baseline DORN and training with data augmentation (i.e., randomly warping images) in terms of generalization performance without requiring data augmentation, hence resulting in shorter training time.
\begin{table}[ht]
\caption{Performance of surface normal estimation on Azure Kinect dataset}
\vspace{-5mm}
\label{tab:surface_normal_azure}
\begin{center}
\resizebox{\columnwidth}{!}{
\begin{tabular}{|c|cccccc|}
\hline
Gravity-aligned frames & MAD$\downarrow$   & Median$\downarrow$ & RMSE$\downarrow$ & 11.25\degree$\uparrow$ & 22.5\degree$\uparrow$ & 30.0\degree$\uparrow$ \\
\hline
Vanilla                         & 9.15  & 4.89   & 15.91 & 80.83    & 90.83   & 93.41 \\
Warping Augmentation    & 9.31  & 4.61   & 17.23 & 81.88    & 90.59   & 92.95   \\
Gravity Alignment & \textbf{8.62}  & \textbf{4.26}   & \textbf{15.57} & \textbf{82.03}    & \textbf{91.17}   & \textbf{93.68}   \\
\hline
\hline
Gravity-nonaligned frames & MAD$\downarrow$   & Median$\downarrow$ & RMSE$\downarrow$ & 11.25\degree$\uparrow$ & 22.5\degree$\uparrow$ & 30.0\degree$\uparrow$ \\
\hline
Vanilla                         & 14.40 & 6.86   & 23.32 & 65.55    & 81.56   & 85.77   \\
Warping Augmentation     & 11.87 & 5.63   & 20.87 & \textbf{74.91}    & \textbf{86.48}   & 89.67   \\
Gravity Alignment  & \textbf{11.46} & \textbf{5.22}   & \textbf{19.88} & 74.24    & 86.29   & \textbf{89.75} \\
\hline
\hline
All frames & MAD$\downarrow$   & Median$\downarrow$ & RMSE$\downarrow$ & 11.25\degree$\uparrow$ & 22.5\degree$\uparrow$ & 30.0\degree$\uparrow$ \\
\hline
Vanilla                         & 11.37 & 5.41   & 19.31 & 74.35    & 86.48   & 89.95   \\
Warping Augmentation     & 10.62 & 5.09   & 19.19 & 78.31    & 88.48   & 91.27   \\
Gravity Alignment  & \textbf{10.03} & \textbf{4.65}   & \textbf{17.84} & \textbf{78.33}    & \textbf{88.73}   & \textbf{91.74} \\
\hline
\end{tabular}}
\end{center}
\vspace{-5mm}
\end{table}

\begin{table}[ht]
\caption{Performance of depth completion on Azure Kinect dataset}
\vspace{-5mm}
\label{tab:azure}
\begin{center}
\resizebox{\columnwidth}{!}{
\begin{tabular}{|c|c|ccccc|}
\hline
 &                 &                   &                   &   $E(\hat{D}, \delta)$       &                     &  \\ \cline{3-7}
 &	RMSE $\downarrow$ &	1.05 $\uparrow$ &	1.10$\uparrow$  &	1.25$\uparrow$ &	$1.25^2 \uparrow$ &	$1.25^3 \uparrow$ \\
\hline
\hline
	
CSPN-NOISY       & 1.461  & 5.10  &  9.03 & 17.02 & 29.52 & 43.25 \\
CSPN-GT          & 1.465  & 8.10  & 11.20 & 17.93 & 29.49 & 42.69 \\
Ours-SD${}^{\dagger}$  & 0.913 & 11.70 & 22.55 & 47.96 & 74.85 & 88.03 \\
Ours-Enriched 100${}^{\dagger}$ & \textbf{0.814} & \textbf{15.09} & \textbf{28.74} & \textbf{57.93} & \textbf{82.23} & \textbf{91.48} \\
\hline
NeuralRGBD       & 0.717  & 20.03 &	36.25 &	65.25 &	86.28 &	94.59 \\
Ours-ID          & 0.534 &	18.64 &	35.54 &	71.00 &	93.84 &	98.20 \\
Ours-SD          & 0.524 &  19.96 & 37.55 & 73.78 & 94.24 & 98.52 \\
Ours-Enriched 100   & 0.496 &  24.21 & \textbf{44.34} & 79.07 & 95.32 & 98.71 \\
Ours-Enriched 100+g & \textbf{0.490} &  \textbf{24.26} & 44.31 & \textbf{79.23} & \textbf{95.65} & \textbf{98.95} \\
\hline
\end{tabular}
}
\end{center}
\vspace{-5mm}
\end{table}

Table~\ref{tab:azure} presents the depth completion performance of our method, as compared to the NeuralRGBD and CSPN, on the Azure Kinect dataset.
To properly compare with the pretrained NeuralRGBD and CSPN, we separately train our DCN network on ScanNet and NYUv2 (denoted with~${}^{\dagger}$).
%
Since CSPN is trained using points sampled from the ground-truth depth, besides evaluating its performance on the triangulated point cloud (CSPN-NOISY), we also assess its performance when the sparse depth of the same locations is extracted from the ground-truth depth (CSPN-GT).
In addition to the previous results, we also include our depth enrichment method with the surface normal computed using our proposed gravity-aware network (Ours-Enriched 100+g).
Table~\ref{tab:azure} shows that our algorithm significantly outperforms the alternative methods in all metrics for cross-dataset evaluation, thus confirming the generalization capability of depth enrichment approach for the network trained on either NYUv2 or ScanNet.
In addition, it shows that the impact of the depth enrichment method is more significant than the accuracy of the surface normal prediction.
Finally, Fig.~\ref{fig:qualitative} qualitatively illustrates our method for both gravity-aligned and gravity-nonaligned scenes on Azure Kinect dataset.
\subsection{Ablation Studies}
In this section, we study the effect of different inputs on the performance of the network. Our main findings are: 

\noindent \textbf{Surface normal input improves the depth prediction}. Table~\ref{tab:inputs} verifies the effectiveness of the surface normal image as an additional input to the DCN, where we trained and tested a network using only the RGB and sparse depth images on the ScanNet dataset.
This is also in line with the findings of~\cite{qiu2019deeplidar}.
\begin{table}[ht]
\caption{Effect of the surface normal input (ScanNet)} 
\vspace{-7mm}
\label{tab:inputs}
\begin{center}
\resizebox{\columnwidth}{!}{
\begin{tabular}{|c|c|ccccc|}
\hline
 &                 &                   &                   &  $E(\hat{D}, \delta)$      &                     &  \\ \cline{3-7}
 &	RMSE $\downarrow$ & 1.05 $\uparrow$ &	1.10$\uparrow$  &	1.25$\uparrow$ &	$1.25^2 \uparrow$ &	$1.25^3 \uparrow$ \\
\hline
\hline
w/o Normal &	0.342 &	45.65 &	69.38 &	90.14 &	97.57 &	99.38 \\
w/ Normal &	\textbf{0.266} & \textbf{54.09} & \textbf{78.41} & \textbf{94.68} & \textbf{98.91} &	\textbf{99.71} \\
\hline
\end{tabular}
}
\end{center}
\vspace{-5mm}
\end{table}

\noindent \textbf{Higher accuracy normals result in more precise depth}.
We examine the performance of our DCN network using surface-normal input with and without the gravity alignment, denoted as Ours-SD+g and Ours-SD in Table~\ref{tab:betterinputs}, respectively.
In order to highlight the impact of the surface-normal accuracy, we evaluate the networks on the gravity-non-aligned subset of the Azure Kinect dataset, which is shown to have a significant improvement on surface-normal accuracy with the gravity alignment (see Sect.~\ref{ssec:exp_azure}).
Table~\ref{tab:betterinputs} illustrates that with gravity alignment, the depth accuracy increases more as compared to ones evaluated on the entire Azure Kinect dataset (see Sect.~\ref{ssec:exp_azure}).
However, due to the high error in some VI-SLAM point estimates, we notice the depth enrichment with gravity does not outperform other baselines such as the one without enrichment and the one without gravity on RMSE and $\delta=1.25^3$, respectively.
\begin{table}[h]
\caption{Better normal results in better depth \hspace{\textwidth} (Gravity-nonaligned Azure)}
\vspace{-5mm}
\label{tab:betterinputs}
\begin{center}
\resizebox{\columnwidth}{!}{
\begin{tabular}{|c|c|ccccc|}
\hline
 &                 &                   &                   &  $E(\hat{D}, \delta)$      &                     &  \\ \cline{3-7}
 &	RMSE $\downarrow$ & 1.05 $\uparrow$ &	1.10$\uparrow$  &	1.25$\uparrow$ &	$1.25^2 \uparrow$ &	$1.25^3 \uparrow$ \\
\hline
\hline
Ours-SD   &	0.540 & 15.58 & 31.33 & 70.81 & 93.86 & 98.11 \\
Ours-SD+g & 0.514 & 15.94 & 32.55 & 71.94 & 95.06 & \textbf{98.66} \\
Ours-Enriched 100   & \textbf{0.510}	& 20.77 & 39.64 & 76.34 & 94.34 & 98.19 \\
Ours-Enriched 100+g & 0.517 & \textbf{22.05} & \textbf{41.52} & \textbf{78.11} & \textbf{95.41} & 98.65 \\
\hline
\end{tabular}
}
\end{center}
\vspace{-3mm}
\end{table}

\noindent \textbf{Training with depth enrichment inputs degrades cross-dataset performance}. So far, we have employed the depth enrichment method on the networks trained with VI-SLAM's sparse depth input. 
In Table~\ref{table:training_de_inputs}, we examine the performance of our DCN that is  \emph{trained with depth enrichment input}, denoted as Train w/ Enriched X, where X is the number of enriched samples. 
Note that, during the inference time, we use the same amount of enriched samples as in the training for those images that depths can be reconstructed from detected planar surface, otherwise, we use the sparse depth.
Due to this unavailability of plane mask in many images, there is large performance gap between networks trained with input as enriched depth and as sparse depth, suggesting that one should apply the depth enrichment technique only on a pre-trained sparse depth network.
\begin{table}[h]
\caption{Training with depth enrichment input (Azure)}
\vspace{-5mm}
\label{table:training_de_inputs}
\begin{center}
\resizebox{\columnwidth}{!}{
\begin{tabular}{|c|c|ccccc|}
\hline
 &                 &                   &                   &  $E(\hat{D}, \delta)$      &                     &  \\ \cline{3-7}
 &	RMSE $\downarrow$ & 1.05 $\uparrow$ &	1.10$\uparrow$  &	1.25$\uparrow$ &	$1.25^2 \uparrow$ &	$1.25^3 \uparrow$ \\
\hline
\hline
Train w/ Enriched 50   & 0.563 &	17.11 & 32.49 & 65.37 & 92.64 & 98.10 \\
Train w/ Enriched 100 & 0.562 & 17.67 & 33.56 & 68.13 & 93.57 & 98.13 \\
Train w/ Enriched 200   & 0.537 & 18.94 & 35.78 & 70.63 & 92.90 & 97.94 \\
Ours-Enriched 100 & \textbf{0.516} & \textbf{19.43} & \textbf{38.18} & \textbf{75.85} & \textbf{94.81} & \textbf{98.48} \\
\hline
\end{tabular}
}
\end{center}
\vspace{-3mm}
\end{table}
\section{Conclusions and Future Work}
In this paper, we introduced straightforward yet highly effective techniques to improve the generalization performance of sparse-to-dense depth completion by (i) transforming the data to a form that the network is familiar with to produce better surface normal estimates, and (ii) generating an enriched sparse depth image that significantly improves the performance both on similar scenes to the training ones and the generalization datasets collected by different devices.
We thoroughly evaluate multiple configurations of our approach and show the superior performance and generalization ability comparing to other state-of-the-art depth completion methods.
As part of the future work, we plan to incorporate the uncertainty of depth inputs to the network. 
Specifically, we will investigate alternative methods for enriching the sparse depth based on uncertainty instead of random sampling.

\bibliographystyle{IEEEtran}
\bibliography{IEEEabrv,references}

\end{document}